\documentclass[runningheads]{llncs}

\usepackage{times}
\usepackage{graphicx}
\usepackage{blindtext}
\usepackage{amsfonts}
\usepackage[ruled,vlined]{algorithm2e}
\usepackage{url}

\begin{document}


\title{Pose-Selective Max Pooling for Measuring Similarity
} 

\titlerunning{Pose-Selective Max Pooling for Measuring Similarity} 

\authorrunning{Xiang Xiang and Trac D. Tran} 

\author{Xiang Xiang$^1$ and Trac D. Tran$^2$} 
\institute{$^1$Dept. of Computer Science \quad \quad $^2$Dept. of Electrical \& Computer Engineering \\
Johns Hopkins University, 3400 N. Charles St, Baltimore, MD 21218, USA \\ 
\email{xxiang@cs.jhu.edu}
}
\maketitle


\begin{abstract}
    In this paper, we deal with two challenges for measuring the similarity of the subject identities in practical video-based face recognition - the variation of the head pose in uncontrolled environments and the computational expense of processing videos. Since the frame-wise feature mean is unable to characterize the pose diversity among frames, we define and preserve the overall pose diversity and closeness in a video. Then, identity will be the only source of variation across videos since the pose varies even within a single video. Instead of simply using all the frames, we select those faces whose pose point is closest to the centroid of the K-means cluster containing that pose point. Then, we represent a video as a bag of frame-wise deep face features while the number of features has been reduced from hundreds to $K$. Since the video representation can well represent the identity, now we measure the subject similarity between two videos as the max correlation among all possible pairs in the two bags of features. On the official 5,000 video-pairs of the YouTube Face dataset for face verification, our algorithm achieves a comparable performance with VGG-face that averages over deep features of all frames. Other vision tasks can also benefit from the generic idea of employing geometric cues to improve the descriptiveness of deep features.
\end{abstract}

\section{Introduction}

In this paper, we are interested in measuring the similarity of one source of variation among videos such as the subject identity in particular. The motivation of this work is as followed.
Given a face video visually affected by confounding factors such as the identity and the head pose,
we compare it against another video by hopefully only measuring the similarity of the subject identity,
even if the frame-level feature characterizes mixed information.
Indeed, deep features from Convolutional Neural Networks (CNN) trained on face images with identity labels are generally not robust to the variation of the {\bf head pose}, which refers to the face's relative orientation with respect to the camera and is the primary challenge in uncontrolled environments. 
Therefore, the emphasis of this paper is not the deep learning of frame-level features.
Instead, we care about how to improve the video-level representation's descriptiveness which rules out confusing factors (\emph{e.g.}, pose)
and induces the similarity of the factor of interest (\emph{e.g.}, identity).

If we treat the frame-level feature vector of a video as a random vector, we may assume that the highly-correlated feature vectors are identically distributed. When the task is to represent the whole image sequence instead of modeling the temporal dynamics such as the state transition, we may use the sample mean and variance to approximate the true distribution, which is implicitly assumed to be a normal distribution. While this assumption might hold given natural image statistics, it can be untrue for a particular video. Even if the features are Gaussian random vectors, taking the mean makes sense only if the frame-level feature just characterizes the identity. Because there is no variation of the identity in a video by construction. However, even the CNN face features still normally contain both the identity and the pose cues. Surely, the feature mean will still characterize both the identity and the pose. What is even worse, there is no way to decouple the two cues once we take the mean. Instead, if we want the video feature to only represent the subject identity, we had better preserve the overall pose diversity that very likely exists among frames. Disregarding minor factors, the identity will be the only source of variation across videos since pose varies even within a single video. The proposed $K$ frame selection algorithm retains frames that preserve the pose diversity. Based on the selection, we further design an algorithm to compute the identity similarity between two sets of deep face features by pooling the max correlation.

\begin{figure}[!t]
\centering
\includegraphics[scale=0.22]{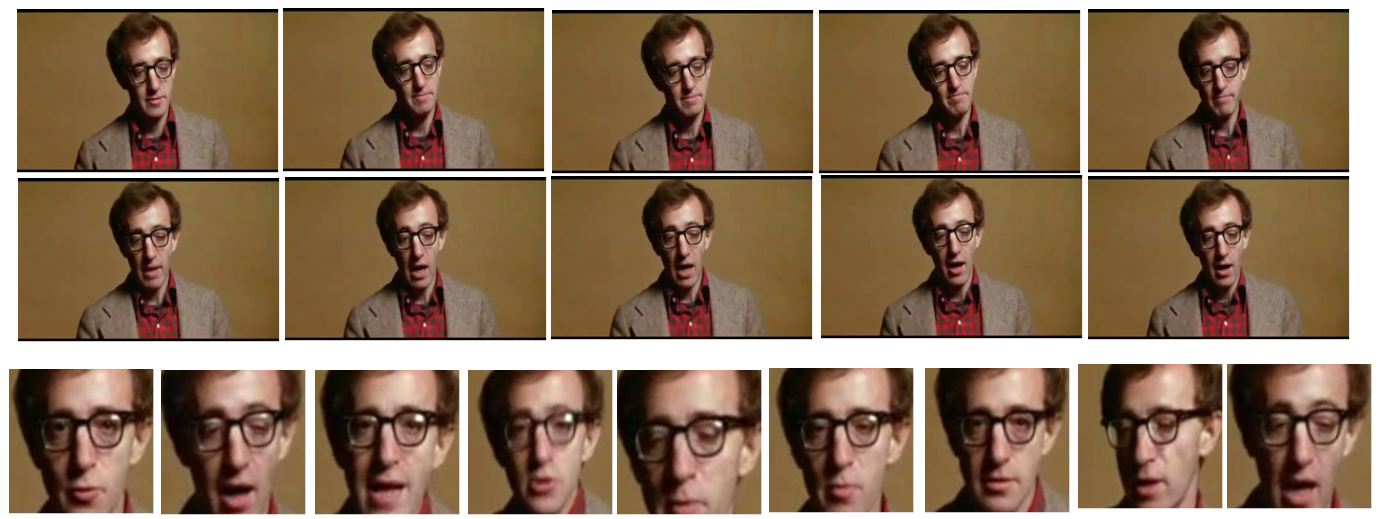}
\vspace{-3mm}
\caption{Example of the chosen key faces. Top row shows the first 10 frames of a 49-frame YTF sequence of Woody Allen, who looks right and down sometimes. And most of the time his face is slightly slanting. Bottom row are 9 frames selected according to the variation of 3D poses. Disclaimer: the source owning this YouTube video allows republishing the face images.}
\label{fig:pose-eg}
\end{figure}

Instead of pooling from all the frames, the $K$ frame selection algorithm is highlighted at firstly the pose quantization via K-means and then the pose selection using the pose distances to the K-means centroids. It reduces the number of features from tens or hundreds to $K$ while still preserving the overall pose diversity, which makes it possible to process a video stream at real time. \noindent Fig. \ref{fig:pose-eg} shows an example sequence in the YouTube Face (YTF) dataset \cite{ytf11}. This algorithm also serves as a way to sample the video frames (to $K$ images). Once the key frames are chosen, we will pool a single number of the similarity between two videos from many pairs of images. The metric to pool from many correlations normally are the mean or the max. Taking the max is essentially finding the nearest neighbor, which is a typical metric for measuring similarity or closeness of two point sets. In our work, the max correlation between two bags of frame-wise CNN features is employed to measure how likely two videos represent the same person. In the end, a video is represented by a single frame's feature which induces nearest neighbors between two sets of selected frames if we treat each frame as a data point. This is essentially a pairwise max pooling process. On the official 5000 video-pairs of YTF dataset \cite{ytf11}, our algorithm achieves a comparable performance with state-of-the-art that averages over deep features of all frames.

\section{Related Works}

The cosine similarity or correlation both are well-defined metrics for measuring the similarity of two images. A simple adaptation to videos will be randomly sampling a frame from each of the video. However, the correlation between two random image samples might characterize cues other than identity (say, the pose similarity). There are existing works on measuring the similarity of two videos using manifold-to-manifold distance \cite{shan15}. However, the straightforward extension of image-based correlation is preferred for its simplicity, such as temporal max or mean pooling \cite{lionel15temporalpool}. The impact of different spatial pooling methods in CNN such as mean pooling, max pooling and $L$-2 pooling, has been discussed in the literature \cite{boureau10icml,boureau10cvpr}. However, pooling over the time domain is not as straightforward as spatial pooling. The frame-wise feature mean is a straightforward video-level representation and yet not a robust statistic. Despite that, temporal mean pooling is conventional to represent a video such as average pooling for video-level representation \cite{ytb8m}, mean encoding for face recognition \cite{template16}, feature averaging for action recognition \cite{LRCN15} and mean pooling for video captioning \cite{LSTM14NAACL}.

Measuring the similarity of subject identity is useful face recognition such as face verification for sure and face identification as well. Face verification is to decide whether two modalities containing faces represent the same person or two different people and thus is important for access control or re-identification tasks. Face identification involves one-to-many similarity, namely a ranked list of one-to-one similarity and thus is important for watch-list surveillance or forensic search tasks. In identification, we gather information about a specific set of individuals to be recognized (\emph{i.e.}, the gallery). At test time, a new image or group of images is presented (\emph{i.e.}, the probe). 

\begin{figure}[!t]
\centering
\vspace{-3mm}
\includegraphics[scale=0.32]{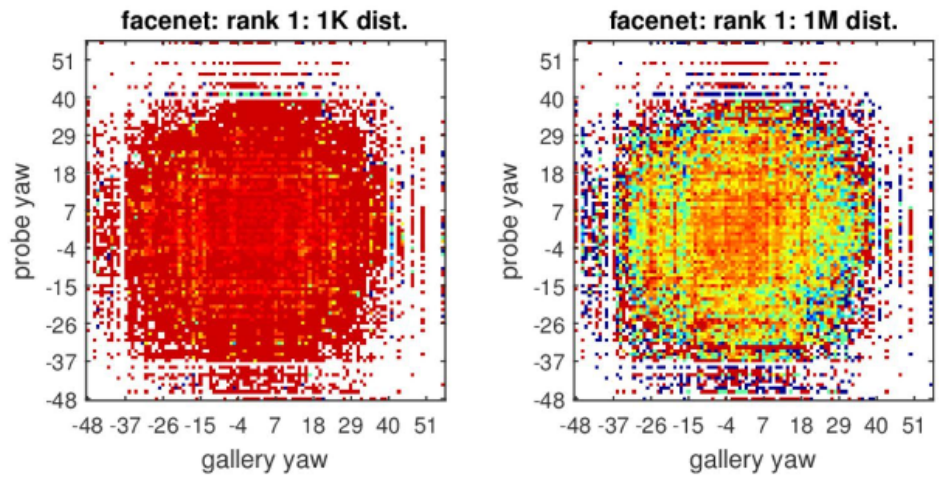}
\caption{Analysis of rank-1 identification under varying poses for Google's FaceNet \cite{facenet15} on the recently established MegaFace 1 million face benchmark \cite{kemelmacher2016megaface}. Yaw is examined as it is the primary variation such as looking left/right inducing a profile. The colors represent identification accuracy going from 0 (blue, none of the true pairs were matched) to 1 (red, all possible combinations of probe and gallery were matched). White color indicates combinations of poses that did not exist in the test set. (a) 1K distractors (people in the gallery yet not in the probe). (b) 1M distractors. This figure is adapted from MegaFace's FaceScrub results.}
\label{fig:facenet}
\end{figure}

In this deep learning era, face verification on a number of benchmarks such as the Labeled Face in the Wild (LFW) dataset \cite{lfw16} has been well solved by DeepFace \cite{deepface14}, DeepID \cite{deepid14}, FaceNet \cite{facenet15} and so on. The Visual Geometry Group at the University of Oxford released their deep face model called VGG-Face Descriptor \cite{vggface15} which also gives a comparable performance on LFW.  However in the real world, pictures are often taken in uncontrolled environment (the so-called in the wild versus in the lab setting). Considering the number of image parameters that were allowed to vary simultaneously, it is logical to consider a divide-and-conquer approach - studying each source of variation separately and keeping all other variations as constants in a control experiment. Such a separation of variables has been widely used in Physics and Biology for multivariate problems. In this data-driven machine learning era, it seems fine to remain all variations in realistic data, given the idea of letting the deep neural networks learn the variations existing in the enormous amount of data. For example, FaceNet \cite{facenet15} trained using a {\bf private} dataset of over 200M subjects is indeed robust to poses, as illustrated in Fig. \ref{fig:facenet}. However, the CNN features from conventional networks suach as DeepFace \cite{deepface14} and VGG-Face \cite{vggface15} are normally not. Moreover, the unconstrained data with fused variations may contain biases towards factors other than identity, since the feature might characterize a mixed information of identity and low-level factors such as pose, illumination, expression, motion and background. For instance, pose similarities normally outweigh subject identity similarities, leading to matching based on pose rather than identity. As a result, it is critical to decouple pose and identity. If the facial expression confuses the identity as well, it is also necessary to decouple them too. In the paper, the face expression is not considered as it is minor compared with pose. Similarly, if we want to measure the similarity of the face expression, we need to decouple it from the identity. For example in \cite{xiang2015hierarchical} for facial expression recognition, one class of training data are formed by face videos with the same expression yet across different people.

Moreover, there are many different application scenarios for face verifications. For Web-based applications, verification is conducted by comparing images to images. The images may be of the same person but were taken at different time or under different conditions. Other than the identity, high-level factors such as the age, gender, ethnicity and so on are not considered in this paper as they remain the same in a video. For online face verification, alive video rather than still images is used. More specifically, the existing video-based verification solutions assume that gallery face images are taken under controlled conditions \cite{shan15}. However, gallery is often built uncontrolled. In practice, a camera could take a picture as well as capture a video. When there are more information describing identities in a video than an image, using a fully live video stream will require expensive computational resources. Normally we need video sampling or a temporal sliding window. 

\section{Pose Selection by Diversity-Preserving K-Means} \label{sec:selection}

In this section, we will explain our treatment particularly for real-world images with various head poses such as images in YTF. Many existing methods such as \cite{xiang2015hierarchical} make a certain assumption which holds only when faces are properly aligned.

By construction (say, face tracking by detection), each video contains a single subject. 
Each video is formalised as a set $\mathbb{V} = \{ \mathbf{x}_1, \mathbf{x}_2,..., \mathbf{x}_m\}$ of frames where each frame $\mathbf{x}_i$ contains a face. Given the homography $\mathbf{H}$ and correspondence of facial landmarks, it is entirely possible to estimate the 3D rotation angles (yaw, pitch and roll) for each 2D face frame. Concretely, some head pose estimator $p(\mathbb{V})$ gives a set $\mathbb{P} = \{ \mathbf{p}_1, \mathbf{p}_2,..., \mathbf{p}_m\}$ where $p_i$ is a 3D rotation-angle vector $({\alpha}_{yaw}, {\alpha}_{pitch}, {\alpha}_{roll})$.

After pose estimation, we would like to select key frames with significant head poses. Our intuition is to preserve pose diversity while downsampling the video in the time domain. We learn from Fig. \ref{fig:facenet} of Google's FaceNet that face features learned from a deep CNN trained on identity-labelled data can be invariant to head poses as long as the training inputs for a particular identity class include almost all possible poses. That is also true for other minor source of variations such as illumination, expression, motion, background among others. Then, identity will be the only source of variation across classes since any factor other than identity varies even within a single class.

Without such huge training data as Google has, we instead hope that the testing inputs for a particular identity class include poses as diverse as possible. A straightforward way is to use the full video, which indeed preserves all possible pose variations in that video while computing deep features for all the frames is computationally expensive. Taking representing a line in a 2D coordinate system as an example, we only needs either two parameters such as the intercept and gradient or any two points in that line. Similarly, now our problem becomes to find a compact pose representation of a testing video which involves the following two criteria. 

\begin{figure}[!t]
\centering
\includegraphics[scale=0.47]{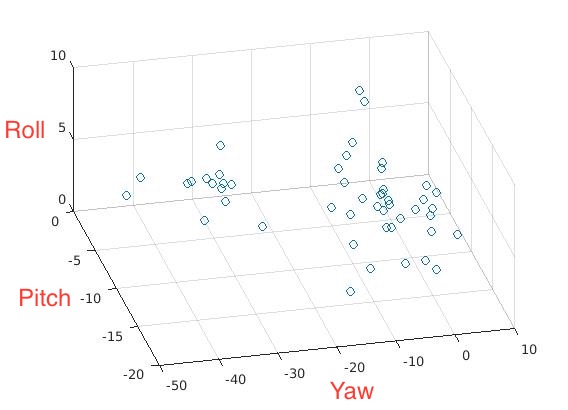}
\caption{An example of 3-D pose space. Shown for the 49-frame Woody Allen sequence in YTF. Three axises represent rotation angles of yaw (looking left or right), pitch (looking up or down) and roll (twisting left or right so that the face is slanting), respectively. The primary variation is the yaw such as turning left/right inducing a profile. Pattern exists in pose distribution - obviously two clusters for this sequence so in extreme case for reducing computation we can set $K=2$.}
\label{fig:pose-pattern}
\end{figure}

First, the pose representation is compact in terms of non-redundancy and closeness. For non-redundancy, we hope to retain as few frames as possible. For pose closeness, we observe from Fig. \ref{fig:pose-pattern} that certain patterns exist in the head pose distribution - close points turn to cluster together. That observation occurs for other sequences as well. As a result, we want to select key frames out of a video by clustering the 3D head poses. The widely-used K-means clustering aims to partition the point set into $K$ subsets so as to minimize the within-cluster Sum of Squared Distances (SSD). If we treat each cluster as a class, we want to minimize the intra-class or within-cluster distance.

Second, the pose representation is representative in terms of diversity (\emph{i.e.}, difference, distance). Intuitively we want to retain the key faces that have poses as different as possible. If we treat each frame's estimated 3D pose as a point, then the approximate polygon formed by selected points should be as close to the true polygon formed by all the points as possible. We measure the diversity using the SSD between any two selected key points (SSD within the set formed by centroids if we use the them as key points). And we want to maximize such a inter-class or between-cluster distance.

Now, we put all criteria together in a single objective. Given a set $\mathbb{P} = \{\mathbf{p}_1, \mathbf{p}_2,..., \mathbf{p}_m\}$ of pose observations, we aim to partition the $m$ observations into $K$ ($\leq m$) {\bf disjoint} subsets $\mathbb{S} = \{\mathbb{S}_1, \mathbb{S}_2, ..., \mathbb{S}_K\}$ so as to minimize the within-cluster SSD as well as maximize the between-cluster SSD while still minimizing the number of clusters:
\begin{equation}\label{eq:kmeans}
\min_{K,\mathbb{S}} \frac{SSD_{within}}{SSD_{between}}
:= \sum_{k=1}^{K} \frac{\sum_{i=1}^{m} \| \mathbf{p}_i - \mathbf{\mu}_k \|^2}{\sum_{j=1,j\ne k}^{K} \| \mathbf{\mu}_j - \mathbf{\mu}_k \|^2}
\end{equation}
where $\mathbf{\mu}_j, \mathbf{\mu}_k$ is the mean of points in $\mathbb{S}_k, \mathbb{S}_k$, respectively. This objective differs from that of K-means only in considering between-cluster distance which makes it a bit similar with multi-class LDA (Linear Discriminant Analysis). However, it is still essentially K-means.
To solve it, we do not really need alternative minimization because that $K$ with a limited number of choices is empirically enumerated by cross validation. Once $K$ is fixed, solving Eqn. \ref{eq:kmeans} follows a similar procedure of multi-class LDA while there is no mixture of classes or clusters because every point is hard-assigned to a single cluster as done in K-means. The subsequent selection of key poses is straightforward (by the distances to K-means centroids). The selected key poses form a subset $\mathbb{P}_{\Omega}$ of $\mathbb{P}$ where $\mathbf{\Omega}$ is a $m$-dimensional $K$-{\bf sparse} impulse vector of binary values 1/0 indicating whether the index is chosen or not, respectively.

    

The selection of frames will follow the index activation vector $\mathbf{\Omega}$ as well. Such a selection reduces the number of images required to represent the face from tens or hundreds to $K$ while preserving the pose diversity which is considered in the formation of clusters. Now we frontalize the chosen faces which is called face alignment or pose correction/normalization. All above operations are summarized in Algorithm \ref{alg:pose}.

\begin{algorithm}
\SetKwInOut{Input}{Input}
\SetKwInOut{Output}{Output}
    \Input{face video $\mathbb{V} = \{ \mathbf{x}_1, \mathbf{x}_2,..., \mathbf{x}_m\}$.}
    \Output{pose-corrected down-sampled face video $\mathbb{V}_{\Omega}^c = \{ \mathbf{x}_{(1)}^c, \mathbf{x}_{(2)}^c,..., \mathbf{x}_{(K)}^c\}$.}
    (1) Landmark detection: detect facial landmarks per frame in $\mathbb{V}$ so that correspondence between frames is known. 
    
    (2) Homography estimation: estimate an approximate 3D model (say, homography $\mathbf{H}$) from the sequence of faces in $\mathbb{V}$ with known correspondence from landmarks.
    
    (3) Pose estimation: compute the rotation angles $p_i$ for each frame using landmark correspondence and obtain a set of sequential head poses $\mathbb{P} = \{ \mathbf{p}_1, \mathbf{p}_2,...,\mathbf{p}_m\} $.
    
    (4) Pose quantization: cluster $\mathbb{P}$ into $K$ subsets $\mathbb{S}_1, \mathbb{S}_2, ..., \mathbb{S}_K$ by solving Eqn. \ref{eq:kmeans} with estimated pose centroids $\{\mathbf{c}_1,...,\mathbf{c}_K\}$ which might be pseudo pose (non-existing pose).
    
    (5) Pose selection: for each cluster, compute the distances from each pose point $\mathbf{p} \in \mathbb{S}_k$ to the pose centroid $c_k$ and then select the closest pose point to represent the cluster $\mathbb{S}_k$. The selected key poses form a subset $\mathbb{P}_{\mathbf{\Omega}}$ of $\mathbb{P}$ where $\mathbf{\Omega}$ is the index activation vector.
    
    (6) Face selection: follow $\mathbf{\Omega}$ to select the key frames and form a subset $\mathbb{V}_{\Omega} = \{ \mathbf{x}_{(1)}, \mathbf{x}_{(2)}, ..., \mathbf{x}_{(K)} \}$ of $\mathbb{V}$ where $\mathbb{V}_{\Omega} \subset \mathbb{V}$.
    
    (7) Face alignment: Warp the each face in $\mathbb{V}_{\Omega}$ according to $\mathbf{H}$ so that landmarks are fixed to canonical positions. 
    \caption{{\bf $K$ frame selection.}}
    \label{alg:pose}
\end{algorithm}

\noindent Note that not all landmarks can be perfectly aligned. Priority is given to salient ones such as the eye center and corners, the nose tip, the mouth corners and the chin. Other properties such as symmetry are also preserved. For example, we mirror the detected eye horizontally. However, a profile will not be frontalized.

\section{Pooling Max Correlation for Measuring Similarity}
In this section, we explain our max correlation guided pooling from a set of deep face features and verify whether the selected key frames are able to well represent identity regardless of pose variation.
 
After face alignment, some feature descriptor, a function $f(\cdot)$, maps each corrected frame $\mathbf{x}_{(i)}^c$ to a $d \times 1$ feature vector $f(\mathbf{x}_{(i)}^c) \in \mathbb{R}^d$ with dimensionality $d$ and unit Euclidean norm. Then the video is represented as a bag of normalized frame-wise CNN features $\mathbb{X} := \{\mathbf{f}_1, \mathbf{f}_2, ..., \mathbf{f}_{K} \}$ $ := \{f(\mathbf{x}_{(1)}^c),f(\mathbf{x}_{(2)}^c),...,f(\mathbf{x}_{(K)}^c)\}$. We can also arrange the feature vectors column by column to form a matrix $\mathbf{X} = \big[\mathbf{f}_1|\mathbf{f}_2|...|\mathbf{f}_K \big]$. For example, the VGG-face network ~\cite{vggface15} has been verified to be able to produce features well representing the identity information. It has 24 layers including several stacked convolution-pooling layer, 2 fully-connected layer and one softmax layer. Since the model was trained for face identification purpose with respect to 2,622 identities, we use the output of the second last fully-connected layer as the feature descriptor, which returns a 4,096-dim feature vector for each input face. 

\begin{figure}[!t]
\centering
\includegraphics[scale=0.27]{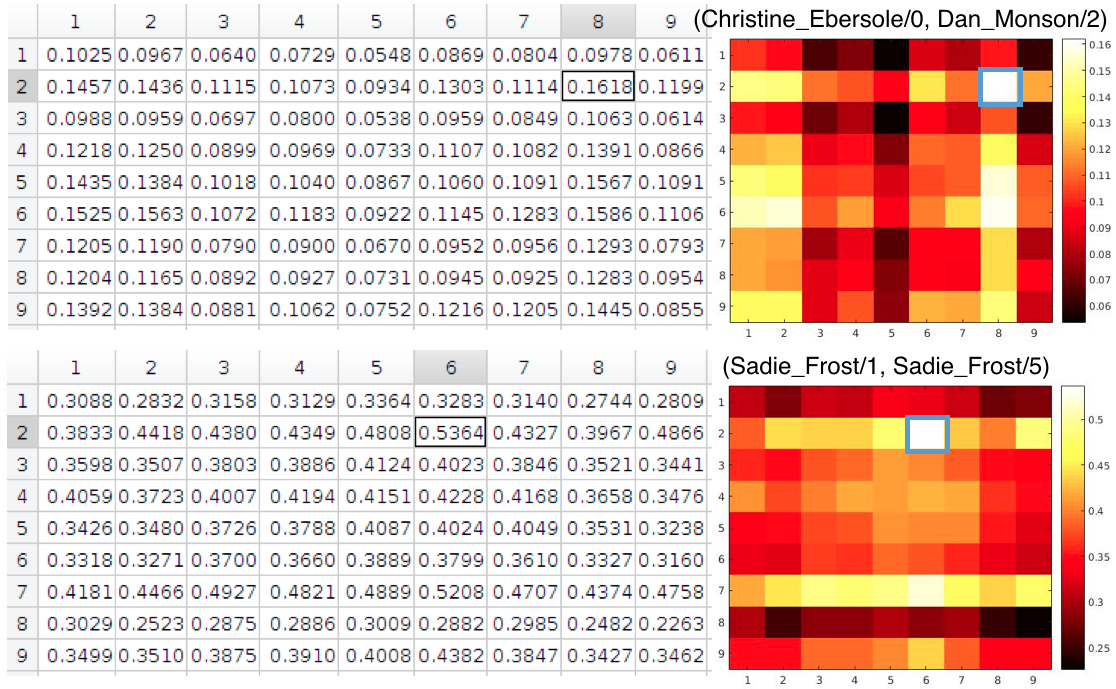}
\caption{Max pooling from the correlation matrix with each axis coordinates the time step in one video. Top row gives an example of different subjects while the bottom row shows that of the same person. Max responses are highlighted by boxes. Faces not shown due to copyright consideration.}
\label{fig:corr}
\end{figure}

Given a pair of videos $(\mathbb{V}_a, \mathbb{V}_b)$ of subject $a$ and $b$ respectively, we want to measure the similarity between $a$ and $b$. Since we claim the proposed bag of CNN features can well represent the identity, instead we will measure the similarity between two sets of CNN features $Sim(\mathbb{X}_a, \mathbb{X}_b)$ which is defined as the max correlation among all possible pairs of CNN features, namely the max element in the correlation matrix (see Fig. \ref{fig:corr}):

\begin{equation}\label{eq:similarity}
Sim(\mathbb{X}_a,\mathbb{X}_b) := \max_{n_a,n_b} ({\mathbf{f}_{n_a}^a}^T \cdot \mathbf{f}_{n_b}^b) = \max \big( ({\mathbf{X}_a}^T {\mathbf{X}_a})(:) \big)
\end{equation}

\noindent where $n_a = 1, 2, ..., K_a$ and $n_b = 1, 2, ..., K_b$. Notably, the notation $(:)$ indicates all elements in a matrix following the MATLAB convention. Now, instead of comparing $m_a \times m_b$ pairs, with Sec. \ref{sec:selection} we only need to compute $K_a \times K_b$ correlations, from which we further pool a single ($1 \times 1$) number as the similarity measure. In the time domain, it also serves as pushing from $K$ images to just $1$ image. The metric can be the mean, median, max or the majority from a histogram while the mean and max are more widely-used. The insight of not taking the mean is that a frame highly correlated with another video usually does not appear twice in a temporal sliding window. If we plot the two bags of features in the common feature space, a similarity is essentially the closeness between the two sets of points. If the two sets are non-overlapping, one measure of the closeness between two points sets is the distance between nearest neighbors, which is essentially pooling the max correlation. Similar with spatial pooling for invariance, taking the max from the correlation matrix shown in Fig. \ref{fig:corr} preserves the temporal invariance that the largest correlation can appear at any time step among the selected frames. Since the identity is consistent in one video, we can claim two videos contain a similar person as long as one pair of frames from each video are highly correlated. The computation of two videos' identity similarity is summarized in Algorithm \ref{alg:sim}.

\vspace{-5mm}

\begin{algorithm}
\SetKwInOut{Input}{Input}
\SetKwInOut{Output}{Output}
    \Input{A pair of face videos $\mathbb{V}_a$ and $\mathbb{V}_b$.}
    \Output{The similarity score $Sim(\mathbb{X}_a,\mathbb{X}_b)$ of their subject identity.}
    \caption{{\bf Video-based identity similarity measurement.}}
    (1) Face selection and alignment: run Algorithm \ref{alg:pose} for each video to obtain key frames with faces aligned.
    
    (2) Deep video representation: generate deep face features of the key frames to obtain two sets of features $\mathbb{X}_a$ and $\mathbb{X}_b$.
    
    (3) Pooling max correlation: compute similarity $Sim(\mathbb{X}_a,\mathbb{X}_b)$ according ro Eqn. \ref{eq:similarity}.
    \label{alg:sim}
\end{algorithm}

\section{Experiments}

\subsection{Implementation}
We develop the programs using resources\footnote{\url{http://opencv.org/}, \url{http://dlib.net/} and \url{http://www.robots.ox.ac.uk/~vgg/software/vgg_face/}, respectively.} such as OpenCV, DLib and VGG-Face.

\begin{itemize}
\item Face detection: frame-by-frame detection using DLib's HOG+SVM based detector trained on 3,000 cropped face images from LFW. It works better for faces in the wild than OpenCV's cascaded haar-like+boosting based (Viola-Jones) detector.
 
\item Facial landmarking: DLib's landmark model trained via regression tree ensemble.

\item Head pose estimation: OpenCV's solvePnP recovering 3D coordinates from 2D coordinates using Direct Linear Transform + Levenberg-Marquardt optimization. 

\item Face alignment: OpenCV's warpAffine by affine-warping to center eyes and mouth.

\item Deep face representation \footnote{Codes are available at \url{https://github.com/eglxiang/vgg_face}}: second last layer output (4,096-dim) of VGG-Face \cite{vggface15} using Caffe \cite{jia2014caffe}. For your conveniece, you may consider using MatConvNet-VLFeat instead of Caffe. VGG-Face has been trained using face images of size 224 $\times$ 224 with the average face image subtracted and then is used for our verification purpose without any re-training. However, such average face subtraction is unavailable and unnecessary given a new inputting image. As a result, we directly input the face image to VGG-Face network without any mean face subtraction.

\end{itemize}

\subsection{Evaluation on video-based face verification}

For video-based face recognition database, EPFL captures 152 people facing web-cam and mobile-phone camera in controlled environments. However, they are frontal faces and thus of no use to us. University of Surrey and University of Queensland capture 295 and 45 subjects under various various well-quantized poses in controlled environments, respectively. Since the poses are well quantized, we can hardly verify our pose quantization and selection algorithm on them. McGill and NICTA capture 60 videos of 60 subjects and 48 surveillance videos of 29 subjects in uncontrolled environments, respectively. However, the database size are way too small. 
YouTube Faces (YTF) dataset (YTF) and India Mvie Face Database (IMFDB) collect 3,425 videos of 1,595 people and 100 videos of 100 actors in uncontrolled environments, respectively. There are quite a few existing work verified on IMFDB. As a result, the YTF dataset \footnote{Dataset is available at \url{http://www.cs.tau.ac.il/~wolf/ytfaces/}} \cite{ytf11} is chosen to verify the proposed video-based similarity measure for face verification. YTF was built by using the 5,749 names of subjects included in the LFW dataset \cite{lfw16} to search YouTube for videos of these same individuals. Then, a screening process reduced the original set of videos from the 18,899 of 3,345 subjects to 3,425 videos of 1,595 subjects.

\begin{figure}[!t]
\centering
\includegraphics[scale=0.5]{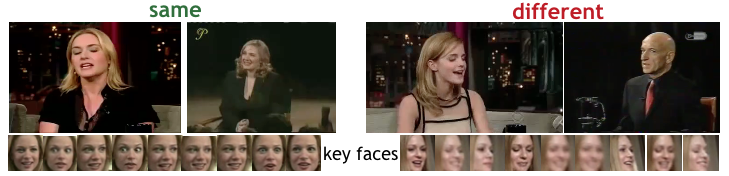}
\caption{Examples of YFT video-pairs. Instead of using the full video in the top row, we choose key faces in the bottom row. Disclaimer: this figure is adapted from VGG-face's presentation (see also \protect\url{http://www.robots.ox.ac.uk/~vgg/publications/2015/Parkhi15/presentation.pptx}) and follows VGG-face's republishing permission.}
\label{fig:ytf-eg}
\end{figure}

\begin{figure}
\centering
\includegraphics[scale=0.26]{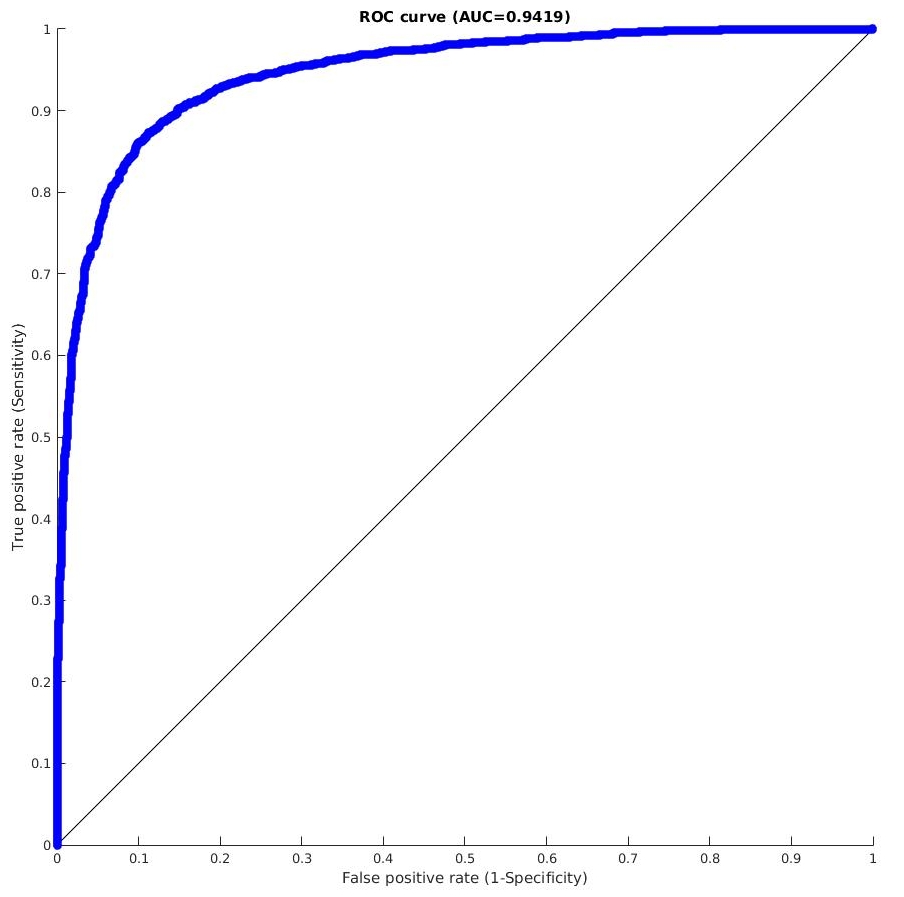}
\caption{ROC curve of running our algorithm on the YTF initial official list of 5,000 pairs.}
\label{fig:roc}
\end{figure}

\begin{figure}
\centering
\includegraphics[scale=0.26]{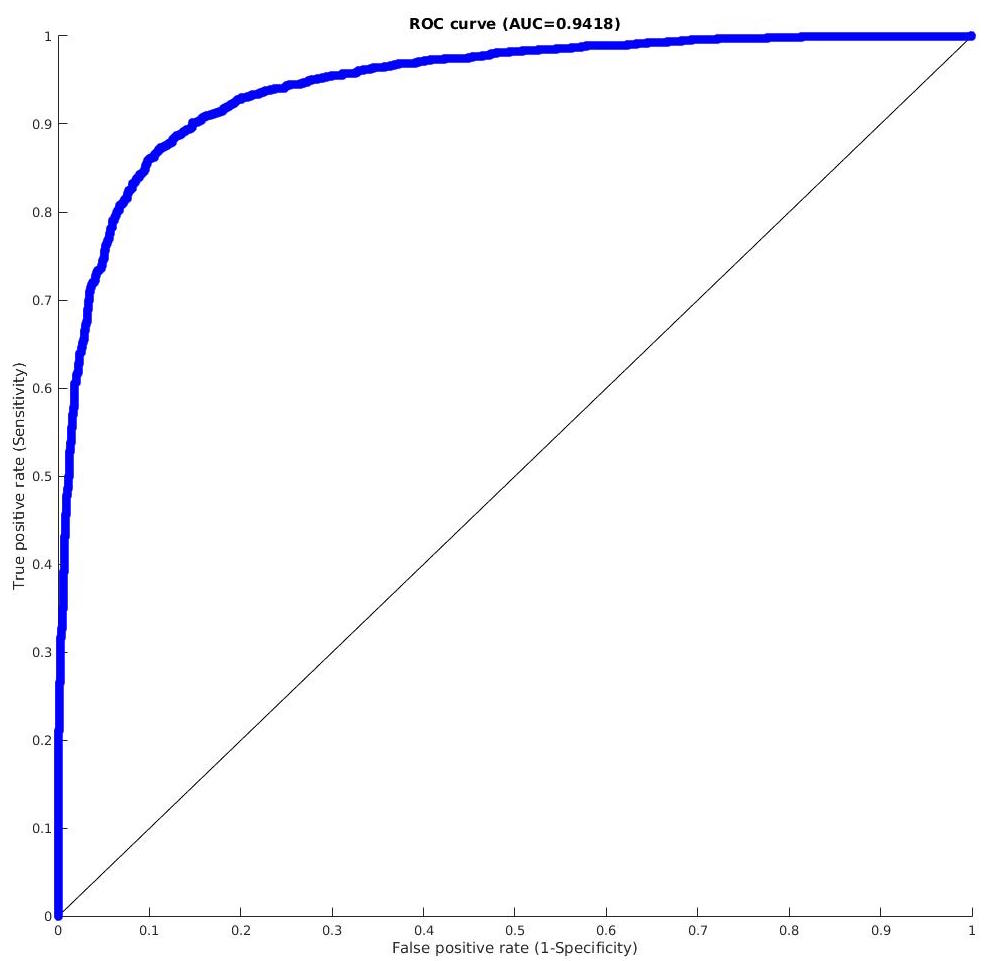}
\caption{ROC curve of running our algorithm on the YTF corrected official list of 4,999 video pairs.}
\label{fig:roc-updated}
\end{figure}

In the same way with LFW, the creator of YTF provides an initial official list of 5,000 video pairs with ground truth (same person or not as shown in Fig. \ref{fig:ytf-eg}). Our experiments can be replicated by following our tutorial \footnote{Codes with a tutorial at \url{https://github.com/eglxiang/ytf}}. $K=9$ turns to be averagely the best for the YTF dataset. Fig. \ref{fig:roc} presents the Receiver Operating Characteristic (ROC) curve obtained after we compute the 5,000 video-video similarity scores. One way to look at a ROC curve is to first fix the level of false positive rate that we can bear (say, 0.1) and then see how high is the true positive rate (say, roughly 0.9). Another way is to see how close the curve towards the top-left corner. Namely, we measure the Area Under the Curve (AUC) and hope it to be as large as possible. In this testing, {\bf the AUC is 0.9419 which is quite close to VGG-Face \cite{vggface15} which uses temporal mean pooling}. However, our selective pooling strategy have much fewer computation credited to the key face selection. We do run cross validations here as we do not have any training.

Later on, the creator of YTF sends a list of errors in the ground-truth label file and provides a corrected list of video pairs with updated ground-truth labels. As a result, we run again the proposed algorithm on the corrected 4,999 video pairs. Fig. \ref{fig:roc-updated} updates the ROC curve with an AUC of 0.9418 which is identical with the result on the initial list.

\section{Conclusion}

In this work, we propose a $K$ frame selection algorithm and an identity similarity measure which employs simple correlations and no learning. It is verified on fast video-based face verification on YTF and achieves comparable performance with VGG-face. Particularly, the selection and pooling significantly reduce the computational expense of processing videos. 
The further verification of the proposed algorithm include the evaluation of video-based face expression recognition. As shown in Fig. 5 of \cite{xiang2015hierarchical}, the assumption of group sparsity might not hold under imperfect alignment. The extended Cohna-Kanade dataset include mostly well-aligned frontal faces and thus is not suitable for our research purpose. Our further experiments are being conducted on the BU-4DFE database\footnote{\url{http://www.cs.binghamton.edu/~lijun/Research/3DFE/3DFE_Analysis.html}} which contains 101 subjects, each one displaying 6 acted facial expressions with moderate head pose variations. A generic problem underneath is variable disentanglement in real data and a take-home message is that employing geometric cues can improve the descriptiveness of deep features. 

\bibliographystyle{splncs03}
\bibliography{lncs}

\begin{thebibliography}{10}
\providecommand{\url}[1]{\texttt{#1}}
\providecommand{\urlprefix}{URL }

\bibitem{ytb8m}
Abu-El-Haija, S., Kothari, N., Lee, J., Natsev, P., Toderici, G., Varadarajan,
  B., Vijayanarasimhan, S.: Youtube-8m: A large-scale video classification
  benchmark. arxiv: 1609.08675  (September 2016)

\bibitem{boureau10cvpr}
Boureau, Y.L., Bach, F., LeCun, Y., Ponce, J.: Learning mid-level features for
  recognition. In: Proceedings of the IEEE Conference on Computer Vision and
  Pattern Recognition (2010)

\bibitem{boureau10icml}
Boureau, Y.L., Ponce, J., LeCun, Y.: A theoretical analysis of feature pooling
  in visual recognition. In: Proceedings of the International Conference on
  Machine Learning (2010)

\bibitem{template16}
Crosswhite, N., Byrne, J., Parkhi, O.M., Stauffer, C., Cao, Q., Zisserman, A.:
  Template adaptation for face verification and identification. arxiv  (April
  2016)

\bibitem{LRCN15}
Donahue, J., Hendricks, L.A., Guadarrama, S., Rohrbach, M., Venugopalan, S.,
  Saenko, K., Darrell, T.: Long-term recurrent convolutional networks for
  visual recognition and description. In: Proceedings of the IEEE Conference on
  Computer Vision and Pattern Recognition. pp. 2625--2634 (2015)

\bibitem{shan15}
Huang, Z., Shan, S., Wang, R., Zhang, H., Lao, S., Kuerban, A., Chen, X.: A
  benchmark and comparative study of video-based face recogni-tion on cox face
  database. IEEE Transaction on Image Processing  24,  5967--5981 (2015)

\bibitem{jia2014caffe}
Jia, Y., Shelhamer, E., Donahue, J., Karayev, S., Long, J., Girshick, R.,
  Guadarrama, S., Darrell, T.: Caffe: Convolutional architecture for fast
  feature embedding. arXiv:1408.5093  (2014)

\bibitem{kemelmacher2016megaface}
Kemelmacher-Shlizerman, I., Seitz, S.M., Miller, D., Brossard, E.: The megaface
  benchmark: 1 million faces for recognition at scale. In: Proceedings of the
  IEEE Conference on Computer Vision and Pattern Recognition (2016)

\bibitem{lfw16}
Learned-Miller, E., Huang, G.B., RoyChowdhury, A., Li, H., Hua, G.: Labeled
  faces in the wild: A survey. Advances in Face Detection and Facial Image
  Analysis pp. 189--248 (2016)

\bibitem{vggface15}
Parkhi, O.M., Vedaldi, A., Zisserman, A.: Deep face recognition. In: British
  Machine Vision Conference (2015)

\bibitem{lionel15temporalpool}
Pigou, L., van~den Oord, A., Dieleman, S., Herreweghe, M.V., Dambre, J.: Beyond
  temporal pooling: Recurrence and temporal convolutions for gesture
  recognition in video. arxiv  (June 2015)

\bibitem{facenet15}
Schroff, F., Kalenichenko, D., Philbin, J.: Facenet: A unified embedding for
  face recognition and clustering. In: Proceedings of the IEEE International
  Conference on Computer Vision (2015)

\bibitem{deepid14}
Sun, Y., Chen, Y., Wang, X., Tang, X.: Deep learning face representation by
  joint identification-verification. In: Advances in Neural Information
  Processing Systems (2014)

\bibitem{deepface14}
Taigman, Y., Yang, M., Ranzato, M., Wolf, L.: Deepface: Closing the gap to
  human-level performance in face verification. In: Proceedings of the IEEE
  International Conference on Computer Vision (2014)

\bibitem{LSTM14NAACL}
Venugopalan, S., Xu, H., Donahue, J., Rohrbach, M., Mooney, R., Saenko, K.:
  Translating videos to natural language using deep recurrent neural networks.
  In: Proceedings of the Annual Conference of the North American Chapter of the
  Association for Computational Linguistics: Human Language Technologies (2014)

\bibitem{ytf11}
Wolf, L., Hassner, T., Maoz, I.: Face recognition in unconstrained videos with
  matched background similarity. In: Proceedings of the IEEE Conference on
  Computer Vision and Pattern Recognition (2011)

\end{thebibliography}

\end{document}